\documentclass[lettersize,journal]{IEEEtran}
\usepackage{algorithm}
\usepackage{array}
\usepackage[caption=false,font=normalsize,labelfont=sf,textfont=sf]{subfig}
\usepackage{stfloats}
\usepackage{url}
\usepackage{verbatim}
\usepackage{graphicx}
\usepackage{cite}
\usepackage{amsmath,amssymb,amsfonts}
\usepackage{algorithmic}
\usepackage{graphicx}
\usepackage{graphicx}
\usepackage{textcomp}
\usepackage{xcolor}
\usepackage{graphicx,color}

\begin{document}
\title{A Predictive Approach for Enhancing Accuracy in Remote Robotic Surgery Using  Informer Model}

\author{Muhammad Hanif Lashari, Shakil Ahmed,~\IEEEmembership{Member,~IEEE}, Wafa Batayneh, and Ashfaq Khokhar,~\IEEEmembership{Fellow,~IEEE}

\vspace{-0.75 cm}
\thanks{All the authors belong to the Department of Electrical \& Computer Engineering, Iowa State University, Ames, USA (e-mails: \{mhanif, shakil, batayneh, ashfaq\}@iastate.edu)}

\thanks{Manuscript received Month DD, YYYY; revised Month DD, YYYY.}}

\markboth{IEEE transactions on cybernetics}%
{Hanif \MakeLowercase{\textit{et al.}}: A Sample Article Using IEEEtran.cls for IEEE Journals}

\maketitle

\begin{abstract}
Precise and real-time estimation of the robotic arm's position on the patient's side is essential for the success of remote robotic surgery in Tactile Internet (TI) environments. This paper presents a prediction model based on the Transformer-based Informer framework for accurate and efficient position estimation. Additionally, combined with a Four-State Hidden Markov Model (4-State HMM) to simulate realistic packet loss scenarios. The proposed approach addresses challenges such as network delays, jitter, and packet loss to ensure reliable and precise operation in remote surgical applications.
The method integrates the optimization problem into the Informer model by embedding constraints such as energy efficiency, smoothness, and robustness into its training process using a differentiable optimization layer. The Informer framework uses features such as ProbSparse attention, attention distilling, and a generative-style decoder to focus on position-critical features while maintaining a low computational complexity of \(O(L \log L)\).
The method is evaluated using the JIGSAWS dataset, achieving a prediction accuracy of over 90\% under various network scenarios. A comparison with models such as TCN, RNN, and LSTM demonstrates the Informer framework’s superior performance in handling position prediction and meeting real-time requirements, making it suitable for Tactile Internet-enabled robotic surgery.
\end{abstract}

 \section{Introduction}
\IEEEPARstart{T}{he} Tactile Internet (TI) represents a significant evolution of the Internet, transitioning from traditional data exchange to enabling real-time haptic communications and control over networks.
TI introduces new possibilities in several areas, including remote robotic surgery, which depends on real-time touch feedback and accuracy\cite{b1}. 
For these applications, reliability should be high, and extremely low latency is required, with end-to-end delays less than 1 millisecond ~\cite{b2}.
Remote robotic surgery will allow surgeons to perform surgical procedures such as Incision, Knot-tying, Suturing, and Needle-Passing~\cite{b88} over vast distances, breaking geographical barriers. However, the success of these surgical procedures is highly dependent on the accurate and timely transmission of haptic commands and feedback between the Surgeon's Side Manipulator (SSM) and the Patient Side Manipulator (PSM)\cite{b3} or Robotic Surgical System.

Some of the critical challenges in remote robotic surgery include network-induced uncertainties such as delays, jitter, and packet loss \cite{b4}. These factors can disrupt the data packet transmission of haptic commands and feedback between the SSM and the PSM, leading to inaccuracies in the PSM's movements. Moreover, packet loss can cause arm's position data loss. This will make it difficult for the robot on the PSM's side to replicate the SSM's intended actions accurately.
Recent advances in teleoperation systems in\cite{b26} have demonstrated the effectiveness of integrating active vision and pose estimation techniques for precise and stable robotic control. These methods highlight the need to address the challenges of real-time position estimation in remote robotic surgery.

With the advent of ultra-responsive connectivity provided by technologies like 5G, the development of the TI has gained significant momentum, especially for applications requiring real-time precision, such as remote robotic surgery. Although advances in network infrastructure have substantially reduced latency, challenges such as packet loss and jitter persist due to physical and environmental limitations \cite{b5}. Integrating prediction-based systems, such as the Informer model, can anticipate and compensate for network-induced uncertainties.
Traditional methods for mitigating these network issues often involve retransmission strategies, which cannot be used in time-critical applications like surgery due to the added latency~\cite{b6}.
Therefore, there is a pressing need for advanced prediction models that can accurately estimate the PSM's position in real time despite challenges posed by network imperfections.

In this research, we presents a new method that utilizes the Informer framework~\cite{b7}, a cutting-edge transformer-based model for long sequence time-series forecasting, to improve position estimation of the PSM in remote robotic surgery. First, we integrated a 4-state HMM to simulate the network's packet loss conditions realistically. This method effectively addresses network-induced delays, jitter, and packet loss. Next, we integrated network simulation with the Informer model to predict the robotic arms. The Informer model's efficient self-attention mechanism and its ability to handle long sequences make it particularly suitable for this application \cite{b7}.
We validated our approach using the publicly available JHU-ISI Gesture and Skill Assessment Working Set (JIGSAWS) dataset~\cite{b8}. Using this dataset, our project demonstrates that the proposed model achieves over 90\% accuracy in position estimation despite adverse network conditions.

The contributions of this research are as follows.
\begin{itemize}
\item Development of a robust prediction model using the Transformer-based Informer architecture for real-time position estimation in remote robotic surgery while maintaining a computational complexity of \(O(L \log L)\). 
\item Integration of the optimization problem (position estimation with certain constraints) into the Informer model by embedding constraints such as energy efficiency, smoothness, and robustness into its training process using a differentiable optimization layer.
\item Introduction of a 4-state HMM for simulating realistic packet loss scenarios, including random and burst errors, for comprehensive model evaluation.
\item Modification of the Informer’s ProbSparse attention mechanism to prioritize position-critical features, enhancing estimation accuracy without increasing computational complexity.
\item Incorporation of network parameters such as latency, jitter, and packet loss into the Informer model, enabling adaptability to varying network conditions.
\item Evaluation and testing of the proposed framework in real-world network conditions with the JIGSAWS dataset and show how well it works for accurate and efficient position estimation.
\item Comparison of the Informer-based framework with other state-of-the-art deep learning models, showcasing its superior performance and real-time constraints.
\end{itemize}

\section{Related Work}
Recent progress in deep learning models has revolutionized time-series forecasting, particularly for scenarios demanding real-time predictions and accurate estimations. Conventional models such as Long Short-Term Memory (LSTM) networks~\cite{b16} and Gated Recurrent Units (GRU)\cite{b17} have been extensively used for robotic control and position estimation\cite{b19}, using their ability to capture temporal patterns in sequential data. Although these models address challenges such as vanishing and exploding gradients, their prediction accuracy remains limited~\cite{b18}.
To address these limitations, Transformer-based models have emerged as a compelling alternative for time-series forecasting. By removing the reliance on sequential processing, Transformers employ a self-attention mechanism to capture long-range dependencies more proficiently~\cite{b10}. Recent applications of Transformers have demonstrated their effectiveness in diverse forecasting tasks, such as multivariate time-series prediction for energy systems and health monitoring in cyber-physical systems~\cite{b7, b20}.

Despite their advantages, standard Transformers struggle with processing very long sequences due to their quadratic complexity in time and memory~\cite{b7}, which restricts their suitability for real-time tasks like remote robotic surgery. As a result, modifications of the Transformer framework, including Temporal Convolutional Networks (TCN) and Convolutional Self-Attention Networks, have been developed to manage long-sequence data with lower computational costs and enhanced prediction performance~\cite{b21}.
The Informer framework enhances performance by employing a ProbSparse self-attention mechanism, reducing computational complexity from \(O(n^2)\) to \(O(n \log n)\). This development in the Informer model makes it effective for handling long-sequence data in real-time settings~\cite{b7}. Furthermore, the Informer’s generative-style decoder mitigates error accumulation in long-term predictions, enhancing its reliability for tracking positional changes over time~\cite{b9}. These features make the Informer a robust choice for predicting the arm's position of the PSM in remote robotic surgery. In such scenarios, challenges such as packet loss and jitter can disrupt performance, but the Informer’s design ensures stability and precision in PSM position estimation, even under challenging network conditions.

\section{Problem Statement}
Remote robotic surgery is a promising application within the Tactile Internet. For remote surgical procedures to succeed, precise and real-time control of the PSM is essential. The PSM needs to carry out commands from the SSM with accuracy, including details such as position, orientation, linear and angular velocity, and the gripper angle of the surgical tools. However, sending the surgeon’s commands over a network can face several challenges including, but not limited to, packet loss, jitter, and delay. These issues, whether they happen as bursts or randomly, can seriously impact the reliability and accuracy of the PSM’s movements.

\subsection{System Model}
\begin{figure}[ht]
    \centering
    \includegraphics[width=0.5\textwidth]{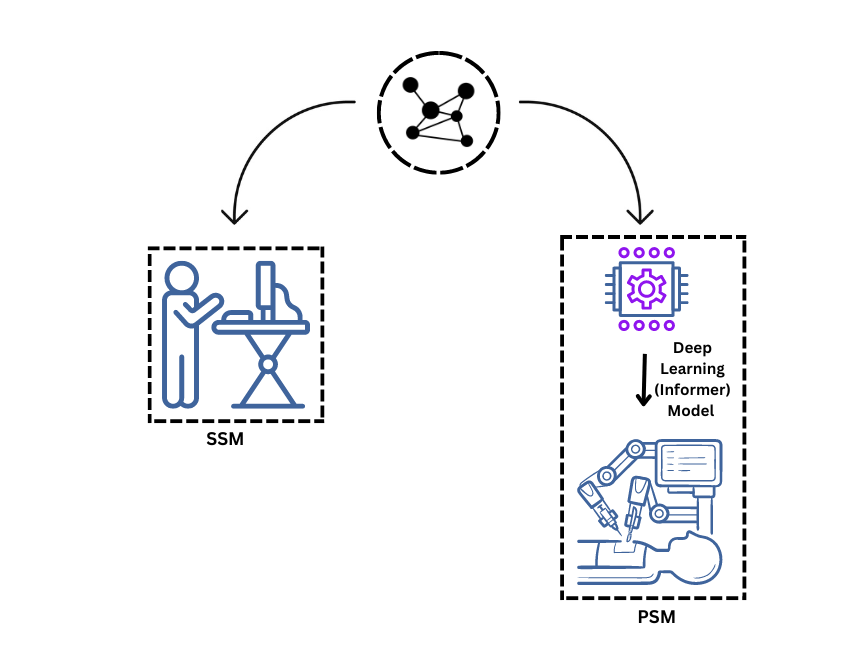}
    \caption{Remote Robotic Surgery Framework Utilizing TI and Informer Model for Enhanced PSM Precision}
    \label{fig:system-model}
\end{figure}

The proposed system model for remote robotic surgery integrates three key domains, as shown in Fig.~\ref{fig:system-model}. The surgeon-side, patient-side, and network domains. Each domain plays a crucial role in the surgical workflow. The surgeon-side domain includes the surgeon console, which captures surgeon gestures and converts them into haptic commands. These commands represent the intended surgical movements, including force, orientation, and kinematic details. The commands are transmitted to the patient's domain, where the PSM executes them highly. The PSM is equipped with a deep learning (Informer model in our case) that estimates and corrects the robot’s arm position in real time. This ensures the surgeon’s movements are accurately replicated despite network-induced losses and delays. The network domain connects these two domains, providing reliable and low-latency communication necessary for seamless execution of the surgery. This system enables precise remote surgeries, addressing distance and network variability challenges using TI technology.

\subsection{System Overview}
Let $\mathbf{x}(t)$ represent the state vector of the PSM at time $t$, which comprises several key operational parameters.
\begin{equation}
\mathbf{x}(t) = \begin{bmatrix}
\mathbf{p}(t) \\
\mathbf{R}(t) \\
\mathbf{v}(t) \\
\mathbf{\omega}(t) \\
\gamma(t)
\end{bmatrix}
\end{equation}
where  $\mathbf{p}(t) = [x(t), y(t), z(t)]^T$ is the 3D position vector of the PSM tool tip.
 $\mathbf{R}(t) \in \mathbb{R}^{3 \times 3}$ is the orientation of the PSM, represented as a rotation matrix.
 $\mathbf{v}(t) = [v_x(t), v_y(t), v_z(t)]^T$ is the linear velocity vector.
$\mathbf{\omega}(t) = [\omega_x(t), \omega_y(t), \omega_z(t)]^T$ is the angular velocity vector.
 $\gamma(t)$ is the gripper angle of the PSM.
The key parameter of interest in this study is the 3D position of the PSM’s robotic arm, represented as \( \mathbf{p}(t) = [x(t), y(t), z(t)]^T \), where \( t \) denotes time. The goal is to ensure that \( \mathbf{p}(t) \), as commanded by the SSM, is accurately executed by the PSM in real time. However, due to network imperfections, the state information \( \mathbf{p}(t) \) received by the PSM can be incomplete or delayed, necessitating the need for a robust prediction model.

\subsection{Network Errors and Challenges}
In this paper, we focus on addressing the critical network-induced errors that impact the performance of the PSM in Tactile Internet-enabled robotic surgery. Specifically, we consider the following types of errors
\begin{itemize}
    \item \textbf{Burst Errors:} These errors occur in clusters, where multiple consecutive packets are lost, leading to significant gaps in transmitted data.
    \item \textbf{Random Errors:} These errors result in the loss of individual packets at random intervals, creating sporadic gaps in the transmitted position data.
\end{itemize}

Both types of errors can significantly degrade the PSM’s ability to accurately replicate the SSM’s commands in real-time. In our previous work \cite{b13}, we addressed random errors using the Kalman Filter (KF) for position estimation, which demonstrated effective compensation. However, the KF struggled to handle burst errors, highlighting the need for a more robust approach. In this study, we focus on mitigating the impact of burst errors using the Informer predictive model to enhance position estimation accuracy under these challenging conditions.

\subsection{Optimization Problem}
We have formulated the position estimation as an optimization problem that aims to improve the accuracy and performance of the PSM. The problem involves multiple parameters, including the tool tip's 3D position\((x, y, z)\), linear velocity, angular velocity, orientation, and gripper angle. However, the primary focus is on minimizing the state estimation error of the 3D position. Considering the complexity introduced by network-induced uncertainties, such as packet loss and jitter, this targeted approach ensures smooth, energy-efficient, and reliable operation under challenging network conditions. The optimization problem is expressed as follows.

\begin{subequations}\label{Eq_op_1}
\begin{align}
&\mathop{\min}_{\hat{\mathbf{p}}(t)} \ \mathbb{E} \left[ \sum_{t=1}^T \left( \|\hat{\mathbf{p}}(t) - \mathbf{p}(t)\|^2 + \alpha E(t) + \beta \Phi(t) \right) \right] \label{Eq_ob_1} \\
&\text{s.t.} \quad \|\hat{\mathbf{p}}(t) - \hat{\mathbf{p}}(t - \Delta t)\| \leq \epsilon_{\text{sync}}, \quad \forall t \label{Eq_op_1_c1} \\
& \hat{\mathbf{p}}(t) \in \mathcal{P}, \quad \mathbf{p}_{\min} \leq \hat{\mathbf{p}}(t) \leq \mathbf{p}_{\max}, \quad \forall t \label{Eq_op_1_c2} \\
& \|\mathbf{v}(t)\| \leq v_{\text{max}}, \quad \|\mathbf{\omega}(t)\| \leq \omega_{\text{max}}, \quad \forall t \label{Eq_op_1_c3} \\
& \|\mathbf{a}(t)\| \leq a_{\text{max}}, \quad \mathbf{a}(t) = \frac{\mathbf{v}(t) - \mathbf{v}(t-1)}{\Delta t}, \quad \forall t \label{Eq_op_1_c4} \\
& \gamma_{\min} \leq \hat{\gamma}(t) \leq \gamma_{\max}, \quad \forall t \label{Eq_op_1_c5} \\
& \sum_{t=1}^T \left( \lambda_v \|\mathbf{v}(t)\|^2 + \lambda_\omega \|\mathbf{\omega}(t)\|^2 \right) \leq E_{\text{max}}. \label{Eq_op_1_c6}
\end{align}
\end{subequations}

The objective function in (\ref{Eq_ob_1}) minimizes the expected weighted sum of three components: state estimation error, energy consumption, and a penalty term for network-induced uncertainties. The state estimation error, represented as \( \|\hat{\mathbf{p}}(t) - \mathbf{p}(t)\|^2 \), ensures that the predicted position \(\hat{\mathbf{p}}(t)\) is as close as possible to the true position \(\mathbf{p}(t)\) at each time step \(t\). The energy consumption \(E(t)\), expressed as \( \lambda_v \|\mathbf{v}(t)\|^2 + \lambda_\omega \|\mathbf{\omega}(t)\|^2 \), accounts for the contributions of linear velocity \(\|\mathbf{v}(t)\|\) and angular velocity \(\|\mathbf{\omega}(t)\|\) to the total power usage. Here, \(\lambda_v\) and \(\lambda_\omega\) are weighting coefficients that balance the relative importance of these velocities. The penalty term \(\Phi(t)\) quantifies the impact of network-induced uncertainties, such as packet loss and jitter, and is modeled using a 4-state HMM. The trade-offs among accuracy, energy efficiency, and robustness are controlled by the weighting coefficients \(\alpha\) and \(\beta\).

The optimization problem is subject to several constraints to ensure operational feasibility. The real-time synchronization constraint in (\ref{Eq_op_1_c1}) enforces that the variation between consecutive predicted positions is below a predefined threshold \(\epsilon_{\text{sync}}\), ensuring that the system operates in real-time. The position feasibility constraint in (\ref{Eq_op_1_c2}) ensures that the predicted position \(\hat{\mathbf{p}}(t)\) lies within the predefined workspace \(\mathcal{P}\), bounded by \(\mathbf{p}_{\min}\) and \(\mathbf{p}_{\max}\).

The velocity limits in (\ref{Eq_op_1_c3}) restrict the linear velocity \(\|\mathbf{v}(t)\|\) and angular velocity \(\|\mathbf{\omega}(t)\|\) to their respective maximum allowable values \(v_{\text{max}}\) and \(\omega_{\text{max}}\). To maintain smooth transitions in position, the smoothness constraint in (\ref{Eq_op_1_c4}) limits the acceleration \(\mathbf{a}(t)\), calculated as \(\mathbf{a}(t) = \frac{\mathbf{v}(t) - \mathbf{v}(t-1)}{\Delta t}\), to an upper bound \(a_{\text{max}}\). The gripper angle constraint in (\ref{Eq_op_1_c5}) ensures that the gripper angle \(\hat{\gamma}(t)\) remains within operational limits \([\gamma_{\min}, \gamma_{\max}]\). Finally, the energy budget constraint in (\ref{Eq_op_1_c6}) ensures that the total energy consumption over the time horizon \(T\) does not exceed \(E_{\text{max}}\).

This optimization problem integrates multiple objectives and constraints to achieve accurate position estimation while maintaining smooth, energy-efficient, and reliable operations under network-induced uncertainties. However, several challenges arise in this context, including ensuring real-time execution, effectively handling both random and burst packet loss, and maintaining computational efficiency. These challenges are critical to achieving precise, smooth, and reliable remote robotic operations. 

To address these challenges, we propose the use of the Transformer-based Informer predictive model. This model is designed to handle complex dependencies and real-time constraints efficiently, providing effective solutions for position estimation. The proposed approach ensures robustness and adaptability under challenging network conditions, improving the overall accuracy and performance of the PSM.

\section{Proposed Model}
In this section, we discuss the proposed approach for solving the problem outlined above. The section is divided into two parts. Part A focuses on modeling network-induced packet loss using a 4-state HMM to simulate realistic loss scenarios, and part B, provides an explanation of the Informer framework, detailing its capabilities and structure.

\subsection{Modeling Packet Loss}
Packet loss in the network is simulated using a 4-state HMM~\cite{b22}, where each state represents a specific network condition, such as. 
\begin{itemize} 
\item \textbf{State 1 ($S_1$):} Successful packet reception during a gap period. 
\item \textbf{State 2 ($S_2$):} Successful packet reception during a burst period. 
\item \textbf{State 3 ($S_3$):} Packet loss during a burst period.
\item \textbf{State 4 ($S_4$):} Packet loss during a gap period. \end{itemize}

Two probabilities govern state transitions. 
\begin{itemize} 
\item \textbf{Burst Density ($P_B$):} Probability of entering or remaining in a burst state. 
\item \textbf{Gap Density ($P_G$):} Probability of entering or remaining in a gap state. 
\end{itemize}

The transition probabilities are represented by the following matrix $\mathbf{T}$.
\begin{equation} 
\mathbf{T} = \begin{bmatrix} 
1 - P_B & P_B & 0 & 0 \\
0 & 1 - P_G & P_G & 0 \\
0 & 0 & 1 - P_G & P_G \\
P_B & 0 & 0 & 1 - P_B \\ 
\end{bmatrix} 
\end{equation}

The 4-state HMM is chosen for its ability to model complex patterns of bursts and random errors, providing a more detailed representation than simpler models like the Gilbert or 2-State HMM~\cite{b23, b24}. This model offers a precise simulation of network conditions by distinguishing between packet loss and successful reception in burst and gap periods, which is essential for high-performance TI environments.

To address the impact of network-induced uncertainties on position estimation, we first simulate packet loss in haptic data. A sequence of haptic data points, representing the position of the SSM over time, is defined as follows. 

\begin{equation} \mathbf{p}(t) = [x(t), y(t), z(t)] \end{equation}
where $t = 1, 2, \dots, T$ and $T$ is the sequence length. Packet loss is simulated by modifying this sequence based on the HMM state at each time step.\begin{itemize} \item For States $S_3$ or $S_4$ (packet loss), the data point is set to zero. \begin{equation} \hat{\mathbf{p}}(t) = [0, 0, 0] \end{equation} \item For States $S_1$ or $S_2$ (packet reception), the data point is preserved. \begin{equation} \hat{\mathbf{p}}(t) = \mathbf{p}(t) \end{equation} \end{itemize}

The resulting sequence \( \hat{\mathbf{p}}(t) \) is processed by the proposed prediction framework to estimate the accurate position of the PSM. 

\subsection{ Informer Model-based Predictive Approch}
\begin{figure}[ht]
\centering
\includegraphics[width=3.5in]{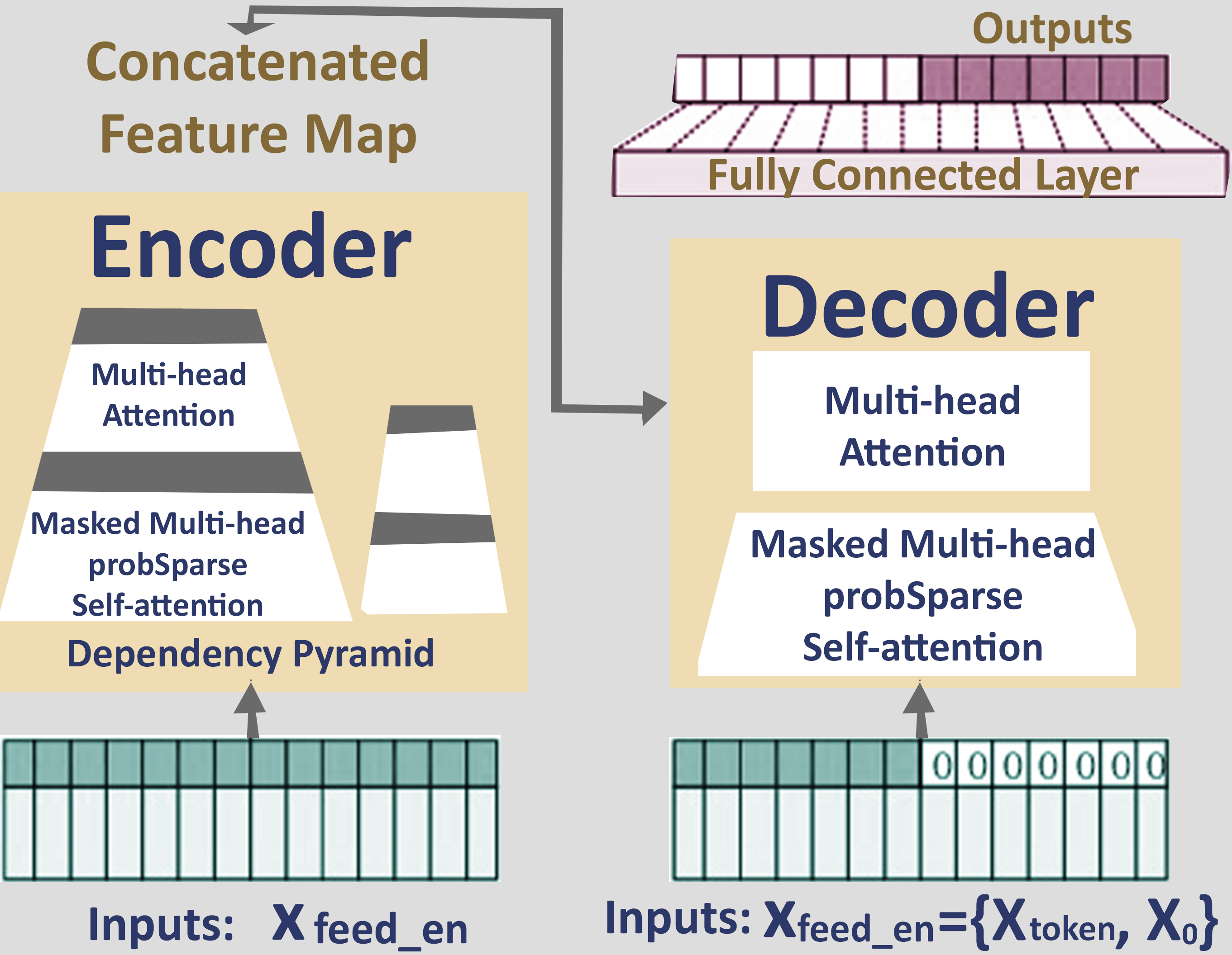}
\caption{Informer Model Encoder-Decoder Framework with ProbSparse Attention Mechanism}
\label{fig:model}
\end{figure}

This study employs the Informer framework, a modified Transformer-based approach, to improve the real-time accuracy of PSM position predictions during remote robotic surgery, as illustrated in Fig.~\ref{fig:model}. Traditional Transformers often struggle with handling long sequences due to their high computational demands and memory usage. The Informer addresses these limitations through key innovations, such as ProbSparse attention, self-attention distilling, and a generative-style decoder. This methodology is adapted from\cite{b7,b9}.

\paragraph{Description of the Informer Model}
This subsection provides a detailed description of each component of the Informer Model and its role in achieving efficient and accurate predictions. 

\paragraph{Optimized Attention Mechanism} In conventional self-attention, as introduced in \cite{b10}, scaled dot-products are computed for queries, keys, and values.
 \begin{equation} 
 A(\mathbf{Q}, \mathbf{K}, \mathbf{V}) = \text{Softmax}\left(\frac{\mathbf{Q}\mathbf{K}^T}{\sqrt{d}}\right) \mathbf{V} \end{equation} where $\mathbf{Q} \in \mathbb{R}^{L_Q \times d}$, $\mathbf{K} \in \mathbb{R}^{L_K \times d}$, and $\mathbf{V} \in \mathbb{R}^{L_V \times d}$. For each query $q_i$, the attention mechanism is defined as. \begin{equation} A(q_i, \mathbf{K}, \mathbf{V}) = \sum_{j} \frac{\exp\left( \frac{q_i k_j^T}{\sqrt{d}} \right)}{\sum_{l} \exp\left( \frac{q_i k_l^T}{\sqrt{d}} \right)} v_j 
 \end{equation}

This method has a computational complexity of $O(L^2)$, making it inefficient for lengthy sequences. To address this, the Informer introduces ProbSparse attention, which reduces computational requirements while retaining accuracy.

\paragraph{Identifying Relevant Queries} To efficiently process extended sequences, the Informer uses a query sparsity measurement based on Kullback-Leibler (KL) divergence. For a given query $q_i$, the attention distribution $p(k_j | q_i)$ is compared with a uniform distribution $u(k_j | q_i) = \frac{1}{L_K}$. The sparsity metric is defined as.  \begin{equation} 
M(q_i, \mathbf{K}) = \ln \left( \sum_{j=1}^{L_K} \exp\left( \frac{q_i k_j^T}{\sqrt{d}} \right) \right) - \frac{1}{L_K} \sum_{j=1}^{L_K} \frac{q_i k_j^T}{\sqrt{d}} 
\end{equation}

This metric identifies the top queries that carry the most critical information.

\paragraph{ProbSparse Attention Mechanism} Using the sparsity metric, the ProbSparse attention mechanism focuses only on significant queries. The updated attention mechanism is defined as. \begin{equation} A(\mathbf{Q}, \mathbf{K}, \mathbf{V}) = \text{Softmax}\left(\frac{\mathbf{Q}\mathbf{K}^T}{\sqrt{d}}\right) \mathbf{V} \end{equation} Here, $\mathbf{Q}$ is a sparse matrix containing the top $u$ queries selected based on $M(q, \mathbf{K})$. The number of key queries $u$ is determined using a factor $c$, set as $u = c \cdot \ln L_Q$, reducing the complexity to $O(L \ln L)$.

\paragraph{Streamlined Self-attention Distilling} The Informer applies self-attention distilling to simplify data processing. Input sequences are condensed layer by layer to emphasize key features and eliminate unnecessary information. The distilling process is described as. \begin{equation} X_{j+1}^t = \text{MaxPool}\left( \text{ELU}(\text{Conv1d}([X_j^t]{AB})) \right) \end{equation} where $[X_j^t]{AB}$ represents the attention block, and Conv1d is a one-dimensional convolutional layer. This reduces memory usage to $O((2 - \epsilon) L \ln L)$.

\paragraph{Efficient Encoder for Long Sequences} The Informer encoder efficiently processes long sequential inputs, balancing memory use and computational efficiency. Input sequences $ \mathbf{X}t $ are transformed into matrices $ \mathbf{X}t^{\text{en}} \in \mathbb{R}^{L_x \times d{\text{model}}} $, with self-attention distilling ensuring only essential features are retained. Inspired by techniques in dilated convolutions~\cite{b11}-\cite{b12}, the transformation from layer $j$ to $j+1$ follows. \begin{equation} X{j+1}^t = \text{MaxPool}\left( \text{ELU}(\text{Conv1d}([X_j^t]_{AB})) \right) \end{equation}

\paragraph{Fast Generative Decoder} The decoder in the Informer model predicts entire sequences in a single pass, ultimately improving speed and reducing cumulative errors. Known tokens $X_{\text{token}}$ and placeholders $X_0$ are used, with masked multi-head attention ensuring predictions remain causal \begin{equation} X_{\text{de}}^t = \text{Concat}(X_{\text{token}}^t, X_0^t) \end{equation}

This design enables fast, precise predictions, making the Informer ideal for real-time applications like remote robotic surgery. We have explained the Informer model and its components. In the next section, we will integrate the optimization problem into the Informer framework to address the challenges in our work.

\section{Integration of Optimization Problem}
Building on the optimization problem formulated in Section III Part B, we now integrate it with the Transformer-based Informer model. The Informer model serves as the foundation for solving this problem by directly embedding the constraints and objectives into its training process. This approach aligns with treating optimization as a differentiable layer, as introduced in \cite{ba1}, and is further guided by insights from optimization principles reviewed in \cite{b25}. By combining the strengths of the Informer model and optimization techniques, this approach enhances predictive accuracy.

\paragraph{Optimization as a Layer} 
The optimization problem, defined to minimize the state estimation error \(\|\hat{\mathbf{p}}(t) - \mathbf{p}(t)\|^2\), while satisfying constraints like energy efficiency and smoothness, is modeled as a differentiable layer. This layer translates as.
\begin{itemize}
    \item Objective Function: Position accuracy and energy efficiency are incorporated as primary and secondary terms in the loss function.
    \begin{equation}
    \mathcal{L}_{\text{total}} = \mathcal{L}_{\text{pos}} + \alpha E(t) + \gamma \|a(t)\|^2 + \beta \Phi(t)
    \end{equation}
    where \(E(t)\) represents energy consumption, \(\|a(t)\|\) ensures smoothness, and \(\Phi(t)\) addresses robustness to network uncertainties.
    \item Constraints: Operational feasibility is maintained through penalties for violating constraints, ensuring the model adheres to real-time requirements.
\end{itemize}

\paragraph{ProbSparse Attention for Position-Critical Features} 
The ProbSparse attention mechanism in the Informer framework is designed to focus computational resources on the most relevant input features, enabling efficient processing of complex data dependencies.
To align the attention mechanism with the optimization problem, the sparsity metric is modified to prioritize position-critical features while deemphasizing secondary features, such as velocity \((\mathbf{v}(t))\) or angular velocity \((\mathbf{\omega}(t))\). 
The updated sparsity metric is defined as.
\begin{equation}
M_{\text{pos}}(q_i, K) = M(q_i, K) + \lambda_1 e_x(t)^T W e_x(t)
\end{equation}
where \(M(q_i, K)\) is the original sparsity metric for attention weights.
 \(e_x(t) = p(t) - \hat{p}(t)\) is the state estimation error for the tooltip position.
\(\lambda_1\) is a scaling factor that prioritizes position estimation error.
\(W\) is the weighting matrix to emphasize certain components of \(x, y, z\).
This modification ensures that the attention mechanism emphasizes features that reduce the position estimation error directly. The original ProbSparse attention mechanism has a complexity of \(O(L \log L)\), where \(L\) is the sequence length. The modification to the sparsity metric adds the position-related term \(e_x(t)^T W e_x(t)\), which involves a matrix-vector multiplication. Since this operation has constant time complexity concerning \(L\), the overall complexity of the attention mechanism remains unchanged at \(O(L \log L)\). This ensures that the attention mechanism remains efficient while prioritizing position-critical features.

\paragraph{Encoder-Guided Constraint Adherence}
The encoder processes input sequences, including corrupted or incomplete position data \(\hat{p}(t)\) arising from packet loss modeled by a 4-state HMM. To ensure that the latent representations align with the constraints of the optimization problem, the encoder incorporates smoothness and energy constraints as regularization terms.
The encoder loss function is designed to minimize.
\begin{equation}
\mathcal{L}_{\text{enc}} = \mathbb{E} \left[ \| \hat{X}_{\text{enc}} - X_{\text{true}} \|^2 + \gamma_1 \| \mathbf{a}(t) \|^2 \right]
\end{equation}
where  \(\hat{X}_{\text{enc}}\) is the latent representation generated by the encoder.
\(X_{\text{true}}\) is the ground truth latent representation.
 \(\mathbf{a}(t) = (\mathbf{v}(t) - \mathbf{v}(t-1)) / \Delta t\) is the acceleration, penalized to enforce smooth movements.
 \(\gamma_1\) is a regularization weight controlling smoothness constraints.

This approach ensures that the encoder generates feasible latent representations. Moreover, incorporating regularization terms, such as the acceleration penalty \(\|a(t)\|^2 = \|v(t) - v(t-1)\|^2 / \Delta t^2\), introduces only constant-time computations per timestep \(t\). These additional operations do not depend on the sequence length \(L\) and, therefore, have negligible impact on complexity. 

\paragraph{Decoder for Real-Time Position Estimation}

The Informer decoder reconstructs the estimated position sequence \((\hat{x}(t), \hat{y}(t), \hat{z}(t))\) by aligning its predictions with the optimization objectives. The decoder’s loss function incorporates the state estimation error as the primary term and energy consumption as a secondary constraint. 
\begin{equation}
\mathcal{L}_{\text{dec}} = \frac{1}{T} \sum_{t=1}^T \| \hat{p}(t) - p_{\text{true}}(t) \|^2 + \delta_1 E(t)
\end{equation}
where \(p_{\text{true}}(t)\) is the true 3D position of the tool tip at time \(t\).
 \(E(t) = \lambda_v \|\mathbf{v}(t)\|^2 + \lambda_\omega \|\mathbf{\omega}(t)\|^2\) represents the energy consumption at time \(t\).
 \(\delta_1\) is a penalty term for energy efficiency constraints.
This formulation ensures that the predicted positions minimize estimation error while adhering to energy constraints, enabling real-time execution. The decoder reconstructs position sequences with a complexity of \(O(L \log L)\), driven by masked self-attention and feedforward operations. Adding energy consumption terms \(E(t) = \lambda_v \|v(t)\|^2 + \lambda_\omega \|\omega(t)\|^2\) in the loss function introduces constant-time operations per timestep \(t\). Since these computations are independent of the sequence length \(L\), the decoder’s complexity remains unchanged.

\paragraph{Incorporating Network Information}

The problem in Section IV can be further enha  nced by explicitly incorporating network parameters such as latency, jitter, and packet loss to better align with the requirements. These factors directly impact real-time performance and robustness of position estimation in the PSM. To address this, the robustness term \(\Phi(t)\) is redefined to include these network-specific metrics. 
\begin{equation}
\Phi(t) \!= [\eta_1 \!\text{PacketLossRate}(t) \!+ \!\eta_2 \text{Latency}(t) \!+\! \eta_3  \text{Jitter}(t)]
\end{equation}
where \(\text{PacketLossRate}(t)\) a proportion of packets lost at time \(t\).
\(\text{Latency}(t)\)is an end-to-end delay of packets at time \(t\).
 \(\text{Jitter}(t)\) is the variability in packet inter-arrival times at time \(t\).
 \(\eta_1, \eta_2, \eta_3\) are weights controlling the contribution of each network parameter to the robustness term.
Incorporating network features as an auxiliary input increases the dimensionality of the input data but does not affect the sequence length \(L\). The complexity of the Informer remains proportional to \(\mathcal{O}(L \log L)\), as the primary cost arises from processing the sequence length, not the dimensionality of the input. Therefore, adding network features has a negligible impact on the overall complexity.
The Informer model is augmented to process network information alongside position data to improve its robustness and adaptability. Network parameters such as predicted latency and jitter are included as auxiliary inputs to the model.
\begin{equation}
\!\!\hat{\mathbf{p}}(t)\! =\! f_{\text{Informer}}(\mathbf{X}_{\text{input}}, \text{PredictedLatency}, \text{PredictedJitter})
\end{equation}
where  \(\mathbf{X}_{\text{input}}\) is the input sequence containing the corrupted or incomplete position data.
 \(\text{PredictedLatency}, \text{PredictedJitter}\) are predicted network conditions at time \(t\), which are included as additional features.
The encoder is modified to account for both position and network conditions. The encoder loss function becomes.
\begin{equation}
\mathcal{L}_{\text{enc}} = \mathbb{E} \left[ \| \hat{X}_{\text{enc}} - X_{\text{true}} \|^2 + \gamma_1 \| a(t) \|^2 + \gamma_2  \text{LatencyPenalty} \right]
\end{equation}
where \(\gamma_2\) weights the penalty for high latency, ensuring the model learns to operate efficiently under varying network conditions.
Network conditions such as latency, jitter, and packet loss are simulated using a 4-state HMM to evaluate the Informer's performance under realistic TI scenarios. These simulations help generate realistic data for training and testing the model. The robustness term \(\Phi(t)\) and the augmented input features allow the Informer to adapt its predictions dynamically. 

\section{Experimental Setup \& Results}
\subsection{Dataset}
The JIGSAWS dataset~\cite{b8} is a publicly available resource containing data from surgical tasks performed using the da Vinci robotic surgical system. This dataset includes synchronized kinematic data, video recordings, and gesture annotations. It was collected during three core surgical tasks, such as knot-tying, suturing, and needle-passing, performed on a bench-top model by eight surgeons with varying skill levels categorized as expert, intermediate, and novice. For this study, 39 trials of the knot-tying task were selected for evaluation.
The kinematic data in the JIGSAWS dataset provides Cartesian positions ($\mathbf{p} \in \mathbb{R}^3$), rotation matrices ($\mathbf{R} \in \mathbb{R}^{3\times3}$), linear velocities ($\mathbf{v} \in \mathbb{R}^3$), rotational velocities ($\omega \in \mathbb{R}^3$), and grasper angles ($\theta$) for both the left and right tools. These features correspond to the PSM and SSM, resulting in 76 attributes sampled at 30 Hz.

\subsection{Simulation Setup}
We conducted the simulations on a system equipped with an Intel Core i7 processor, 32GB of RAM, running Linux OS, and using Python with PyTorch as the primary framework. The experiments were carried out in Jupyter Notebook, which was used for coding and execution. Data preprocessing, model training, and evaluations were performed entirely within this environment. Packet loss patterns were programmatically generated to simulate realistic network conditions, mimicking real Tactile Internet scenarios.

\subsection{Results and Discussion}
\subsubsection{Impact of Packet Loss on Position Estimation}
\begin{figure}[ht]
\centering
\includegraphics[width=3.50in]{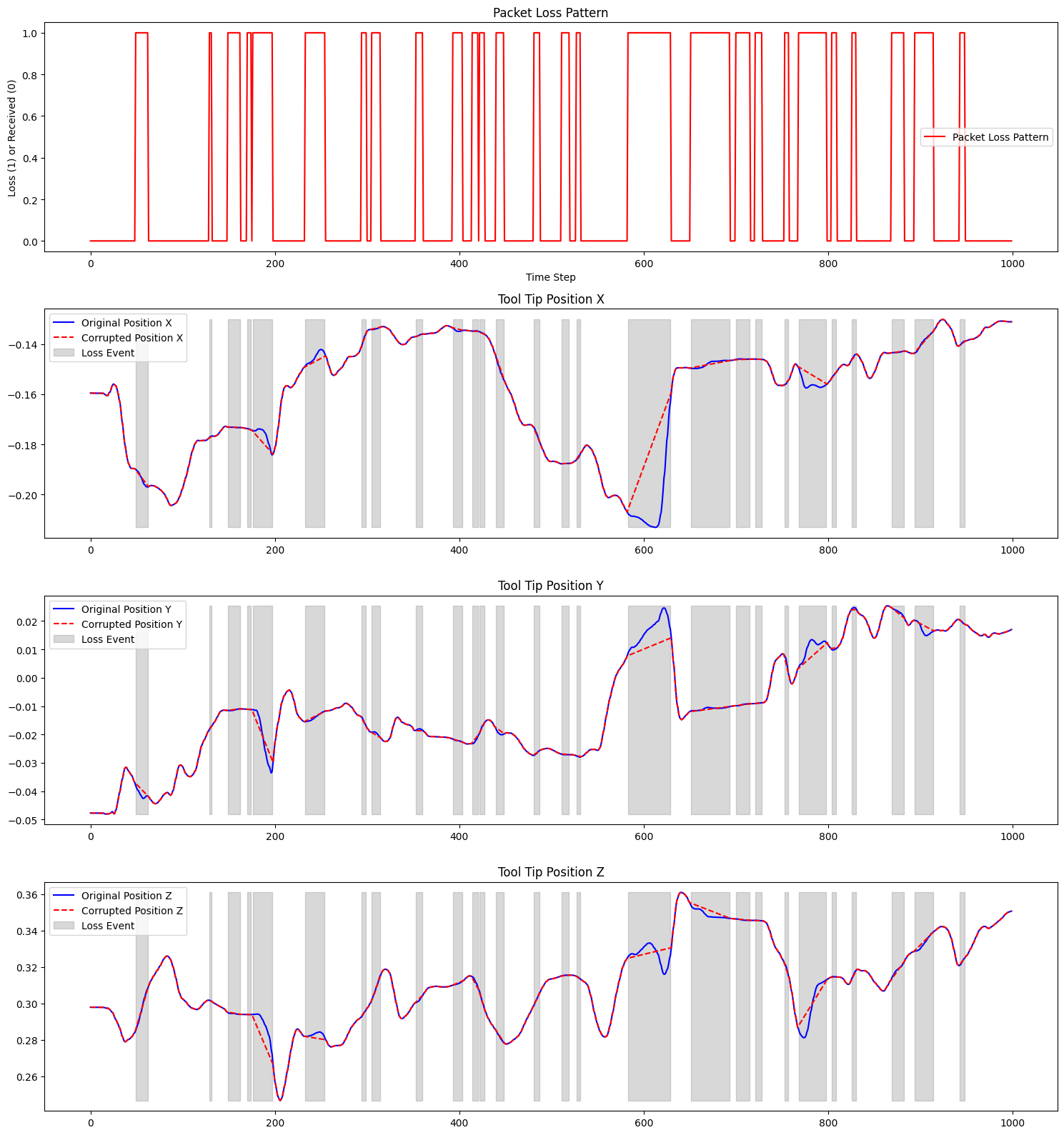}
\caption{The top plot (Part 1) shows the simulated packet loss pattern across 1000 time steps. The following plots (Parts 2-4) display the original and corrupted tool tip position along all three axes, with gray-shaded regions highlighting periods of packet loss.}
    \label{fig:Result1}
\end{figure}

The Fig.~\ref{fig:Result1} part 1, illustrates the packet loss pattern over 1000 time steps. The red spikes indicate the moments when packets were lost (with a value of 1) and successfully received (value of 0). This pattern highlights packet loss's irregular and bursty nature, effectively simulating real-world network conditions.

Figure~\ref{fig:Result1}, parts 2–4, depict the original and corrupted positions of the robotic arm's tool tip along the X, Y, and Z axes under simulated packet loss conditions. The solid blue line represents the ground truth position, while the red dashed line illustrates the corrupted position data resulting from packet loss. The gray areas indicate the time intervals during which packet loss occurs.

\subsubsection{Performance of the Model under Packet Loss}
\begin{figure}[ht]
\centering
\includegraphics[width=3.5in]{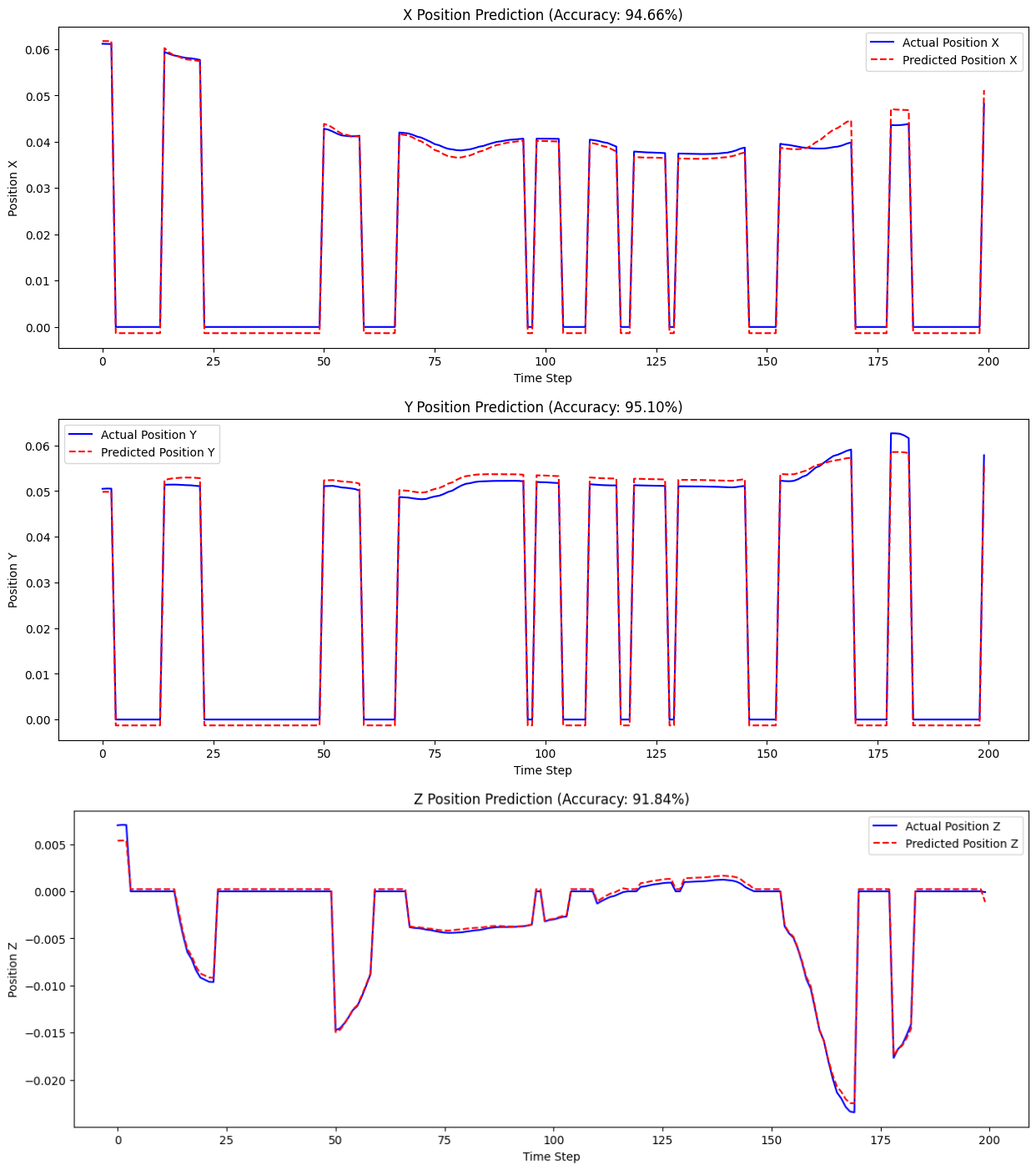}

    \caption{Prediction performance of the Informer model under packet loss for tool tip position in X, Y, and Z axes. Solid and dashed lines represent actual and predicted positions, respectively. The model achieves accuracies of 96.68\%, 95.96\%, and 90.37\% for the X, Y, and Z axes, demonstrating robustness against network-induced packet loss.}
    \label{fig:Result2}
\end{figure}

Fig.~\ref{fig:Result2} shows how the Informer model predicts the robotic arm’s tool tip position along all axes in bursty packet loss. Each plot compares the ground truth and the estimated position for 200 test time steps. 
\begin{itemize}
\item The X position prediction accuracy is 96.68\%.  The model achieves high accuracy in predicting the X-axis position. The predicted position closely follows the actual position, with very few deviations, indicating that the model handles packet loss well for this axis.
\item  Y position prediction accuracy is 95.96\%. Similarly, the model performs effectively in predicting the Y-axis position. The predicted values align almost perfectly with the actual values, except for minor deviations during sharp transitions, demonstrating the robustness of the model.
\item  Z position prediction accuracy is 90.37\%. The Z-axis shows a slightly lower accuracy than the X and Y axes, with some noticeable deviations during time steps where the actual position exhibits rapid changes. However, the overall prediction still captures the trend of position movements, showing that the model can still predict reasonably well in challenging packet loss scenarios.
\end{itemize}
In Table~\ref{tab:performance_metrics}, the Informer model's performance was evaluated under varying network conditions, including different burst densities, gap densities, burst lengths, and gap lengths. 
For comparison, we have evaluated the Informer framework alongside other deep learning models using the same data subset, with the results summarized in Table~\ref{fig:system-model}. The Informer model outperforms TCN, RNN, and LSTM in predicting tool tip positions under packet loss scenarios. Its ProbSparse attention mechanism reduces the computational complexity from \(O(L^2)\) in traditional self-attention to \(O(L \log L)\). TCN, with a complexity of \(O(L \cdot k)\) (where \(k\) is the filter size), struggles with fixed receptive fields, making it less effective in burst error scenarios. RNNs, with a complexity of \(O(L \cdot d^2)\), are hindered by vanishing gradients, which limit their ability to capture temporal correlations. LSTMs address this issue with gating mechanisms but have the same complexity \(O(L \cdot d^2)\) and higher computational costs due to sequential processing. These characteristics make the Informer the most effective model for this task.

\section{Conclusion}

This paper presented a predictive approach using the Transformer-based Informer model to enhance position estimation accuracy in remote robotic surgery. A 4-state HMM was employed to simulate realistic packet loss scenarios, addressing both burst and random loss conditions. The Informer model effectively mitigated network-induced uncertainties, such as jitter and delay, ensuring accurate real-time predictions. Experimental results demonstrated a prediction accuracy exceeding 90\% under diverse network conditions, outperforming traditional models like LSTM and RNN. The integration of constraints such as energy efficiency, smoothness, and robustness further validated the model's suitability for Tactile Internet-enabled surgical applications.

Future work will focus on several key areas. First, the framework can be extended to incorporate multi-objective optimization, balancing position estimation accuracy, latency reduction, and surgical task precision. Second, adaptive mechanisms will be explored to dynamically address varying network conditions, including fluctuating levels of jitter, delay, and packet loss. Lastly, the model’s generalizability will be evaluated through cross-domain validation on diverse surgical datasets and tasks beyond knot-tying, ensuring its applicability to a broader range of robotic-assisted medical procedures.

\begin{table}[ht]
    \centering
    \caption{Comparison of Deep Learning Models for Position Estimation}
    \label{tab:model_comparison}
    \begin{tabular}{|c|c|c|c|}
        \hline
        \textbf{Model} & \textbf{MSE} & \textbf{MAE} & \textbf{RMSE} \\ \hline
        \textbf{Informer} & \textbf{0.0192} & \textbf{0.1082} & \textbf{0.1385} \\ \hline
        TCN & 0.0724 & 0.1313 & 0.2691 \\ \hline
        RNN & 0.1368 & 0.1982 & 0.3699 \\ \hline
        LSTM & 0.1472 & 0.2004 & 0.3837 \\ \hline
    \end{tabular}
\end{table}

\begin{table*}[htbp]
\centering
\caption{Informer Model Performance Metrics at Varying Burst and Gap Densities, Burst Lengths, and Gap Lengths}
\label{tab:performance_metrics}
\begin{tabular}{|c|c|c|c|c|c|c|c|c|c|}
\hline
\textbf{Burst Density} & \textbf{Gap Density} & \textbf{Burst Length} & \textbf{Gap Length} & \textbf{MSE} & \textbf{MAE} & \textbf{RMSE} & \textbf{Accuracy X (\%)} & \textbf{Accuracy Y (\%)} & \textbf{Accuracy Z (\%)} \\ \hline
0.3 & 0.95 & 4  & 8  & 0.0105 & 0.0725 & 0.1027 & 94.27 & 94.25 & 93.40 \\ \hline
0.4 & 0.90 & 5  & 7  & 0.0119 & 0.0771 & 0.1090 & 93.45 & 92.30 & 91.22 \\ \hline
0.5 & 0.85 & 6  & 6  & 0.0116 & 0.0768 & 0.1078 & 92.88 & 91.78 & 90.33 \\ \hline
0.6 & 0.80 & 8  & 5  & 0.0123 & 0.0785 & 0.1109 & 91.33 & 90.22 & 89.12 \\ \hline
0.7 & 0.75 & 10 & 4  & 0.0130 & 0.0792 & 0.1131 & 90.50 & 89.45 & 88.55 \\ \hline
0.8 & 0.70 & 12 & 3  & 0.0136 & 0.0811 & 0.1166 & 89.12 & 88.90 & 87.50 \\ \hline
\end{tabular}
\end{table*}

\section*{Acknowledgment}
The Palmer Department Chair Endowment at Iowa State University partially supported the work in this article.

\begin{IEEEbiography}[{\includegraphics[width=1in,height=1.25in,clip,keepaspectratio]{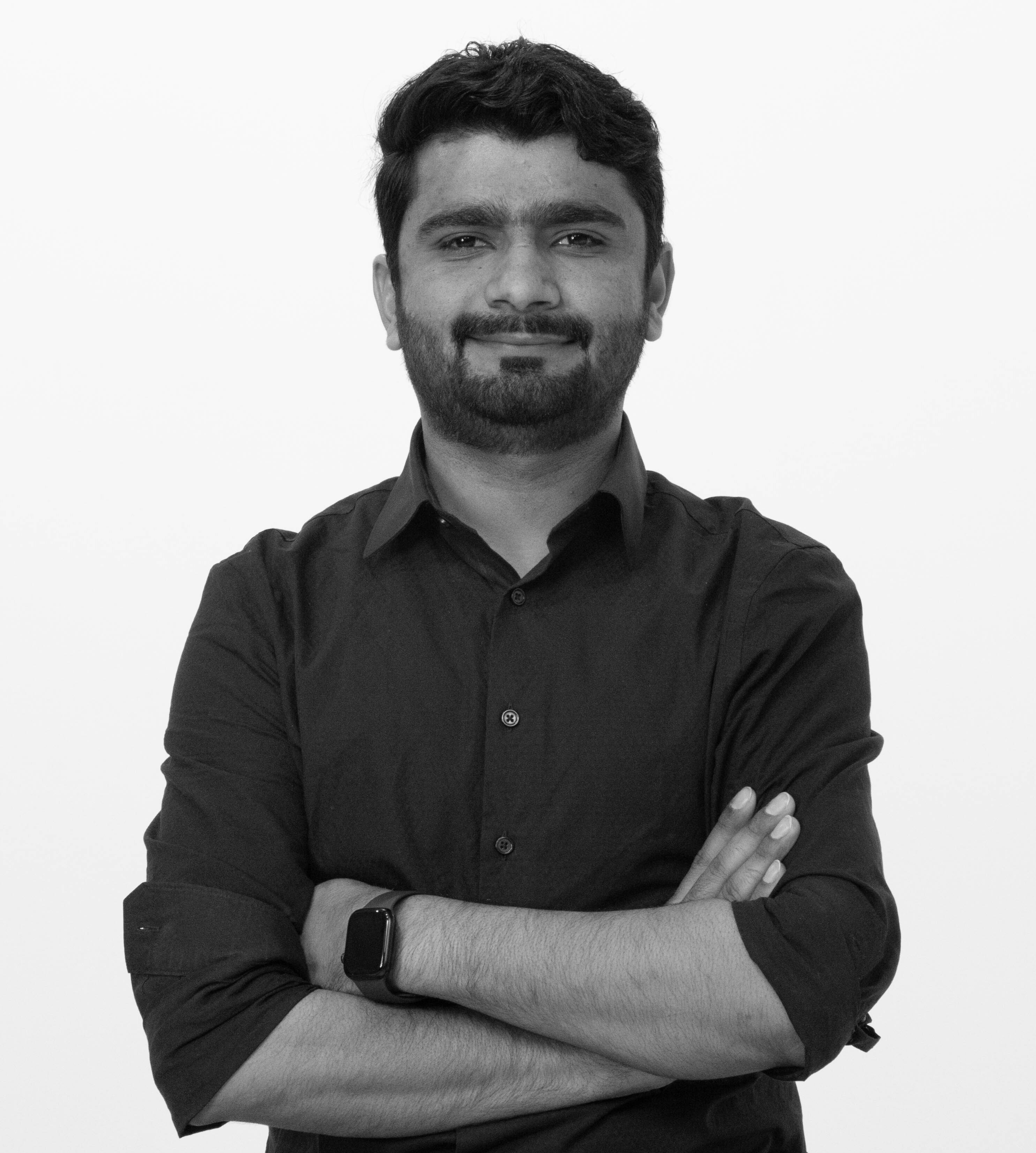}}]
{Muhammad Hanif Lashari}
was born in Sehwan, Sindh, Pakistan. He earned his Bachelor of Engineering (B.E.) in Electronic Engineering in 2016 and his Master of Engineering (M.E.) in Electrical Engineering in 2019 from Mehran University of Engineering \& Technology, Jamshoro, Pakistan. He is currently pursuing his Ph.D. in Computer Engineering at Iowa State University, Ames, Iowa, USA. He has served as a Lecturer at a public sector university in Pakistan and is presently working as a Graduate Research Assistant in the Department of Electrical and Computer Engineering at Iowa State University. His research interests include the Internet of Things, Tactile Internet, Machine Learning, and Deep Learning, with a particular emphasis on enhancing prediction in Tactile Internet applications. He has published three journal and two conference papers and continues to contribute actively to his field. Mr. Lashari is also engaged in various academic activities beyond his immediate research, aiming to broaden his contributions to the field of computer engineering.
\end{IEEEbiography}

\begin{IEEEbiography}[{\includegraphics[width=1in,height=1.25in,clip,keepaspectratio]{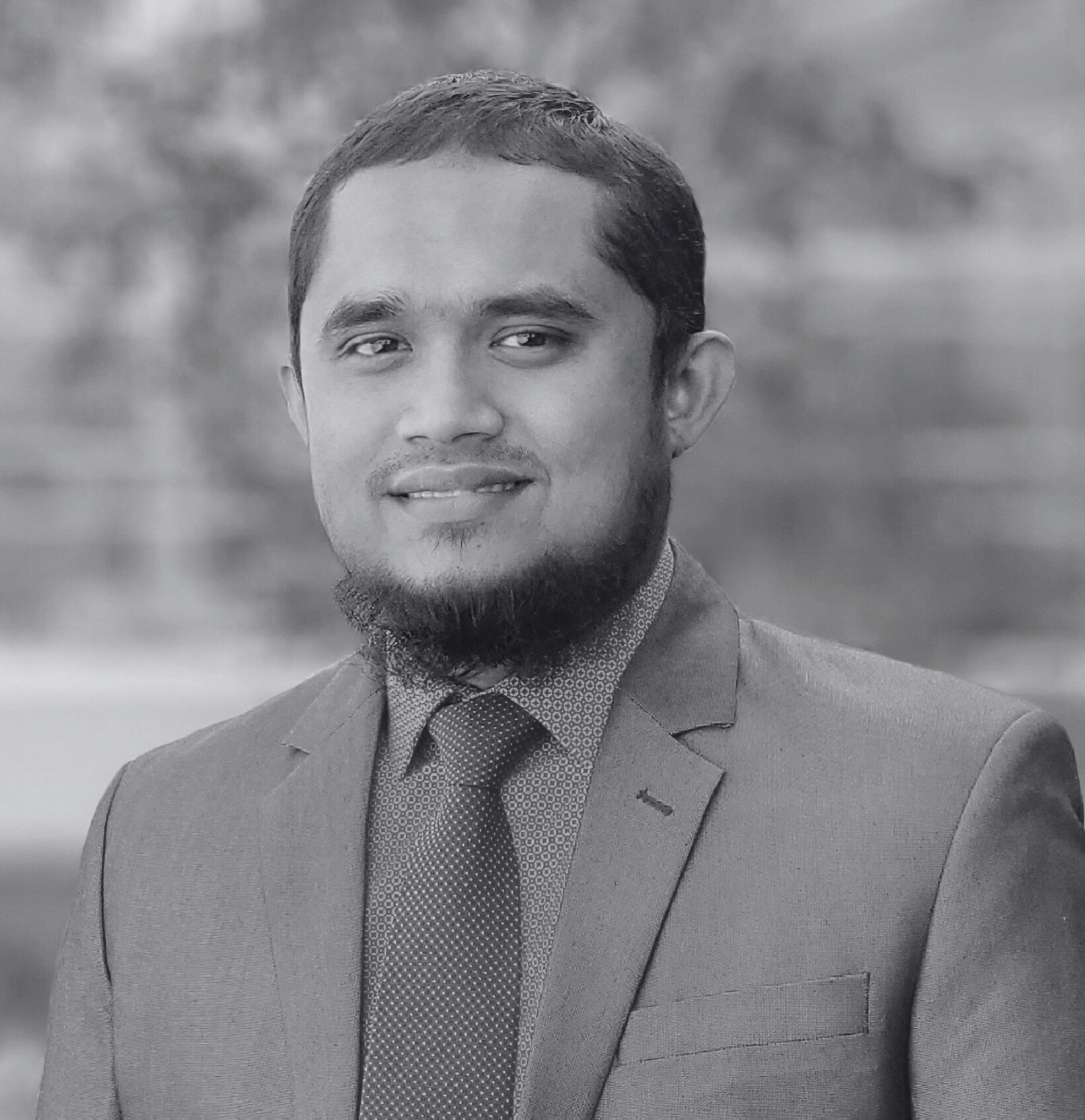}}]
{Shakil Ahmed}~(Member, IEEE)~is an Assistant Teaching Professor in the Department of Electrical and Computer Engineering at Iowa State University. He received his B.S. degree in Electrical and Electronic Engineering from Khulna University of Engineering \& Technology, Bangladesh, in 2014 and his M.S. degree in Electrical Engineering from Utah State University, USA, in 2019. He later earned his Ph.D. in Electrical and Computer Engineering from Iowa State University in a record time of 2 years and 8 months. Ahmed has published numerous research papers in renowned international conferences and journals and received the Best Paper Award at international venues. His research interests include cutting-edge areas such as next-generation wireless communications, GenAI, wireless network design and optimization, unmanned aerial vehicles, physical layer security, Digital Twin for wireless communications, content creation using R.F. signals, and reconfigurable intelligent systems. He is also passionate about engineering education and integrating generative A.I. into learning processes. He has also been a guest editor and reviewer for prestigious journals, including IEEE Transactions on Cognitive Communications and Networking, IEEE Access, IEEE Systems Journal, and IEEE Transactions on Vehicular Technology.
\end{IEEEbiography}

\begin{IEEEbiography}[{\includegraphics[width=1in,height=1.25in,clip,keepaspectratio]{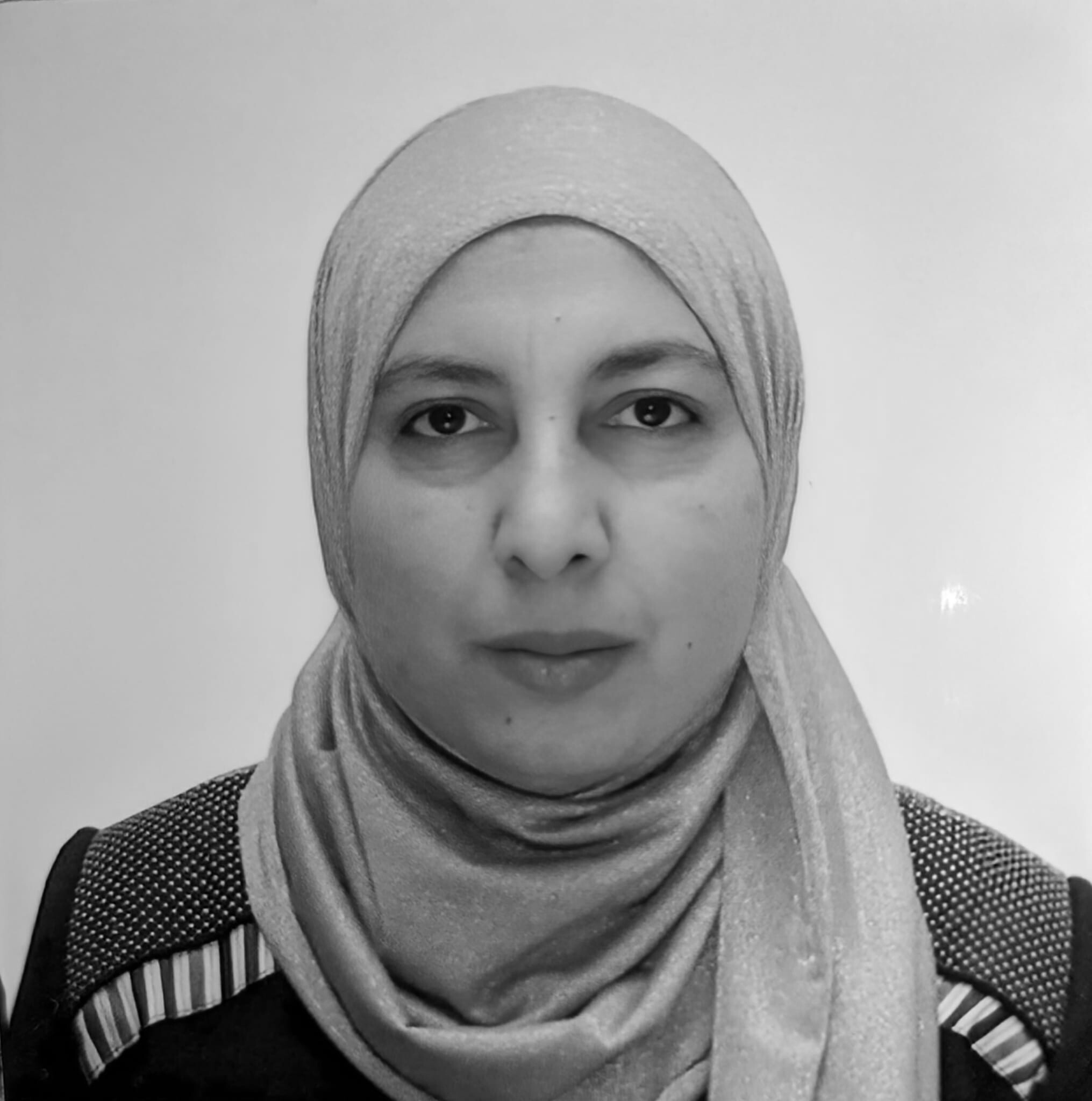}}]
{Wafa Batayneh }was born in Irbid, Jordan. She received her Master's degree and Ph.D. in mechanical engineering from Rensselaer Polytechnic Institute, Troy, NY, USA, in 2005. Her studies focused on Mechatronics systems design.
She has held various positions in academia and research, including summer and fellowship roles, which have contributed to her expertise in robotics and embedded control systems. Currently, she is a Visiting Scholar at the Department of Electrical and Computer Engineering at Iowa State University, Ames, IA, USA, and a Professor at the Mechanical Engineering Department at Jordan University of Science and Technology. She has authored significant publications, including Mechanics of Robotics (Springer, 2010).
Prof. Batayneh is a member of several professional societies. She has received numerous awards for her research and contributions, including national awards.
\end{IEEEbiography}

\begin{IEEEbiography}[{\includegraphics[width=1in,height=1.25in,clip,keepaspectratio]{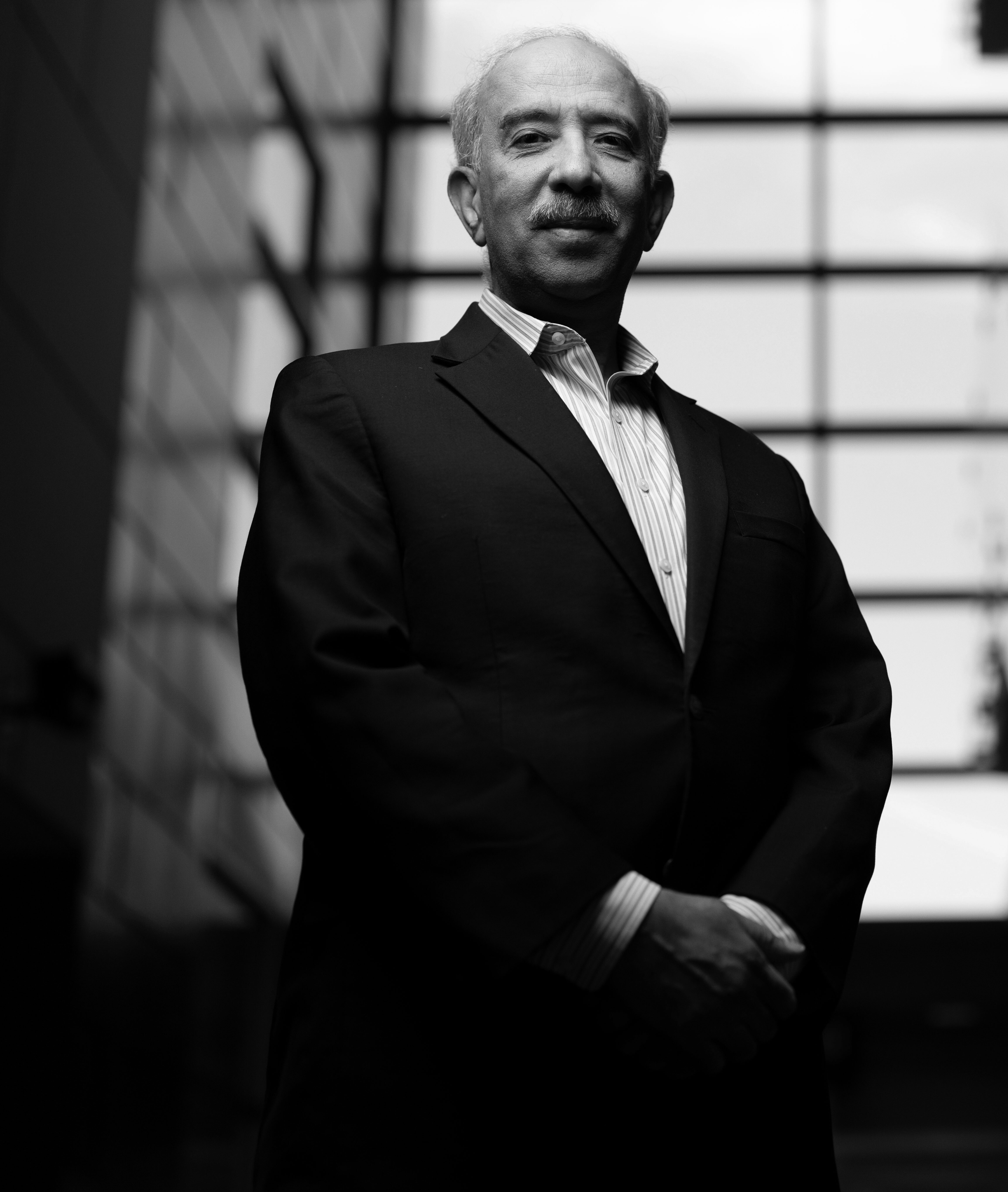}}]
{Ashfaq Khokhar}~(Fellow, IEEE)~currently serves as the Palmer Department Chair of Electrical and Computer Engineering at Iowa State University, a position he has held since January 2017. Previously, he was the Chair of the Department of Electrical and Computer Engineering at the Illinois Institute of Technology from 2013 to 2016. Before that, Khokhar served as a Professor and Director of Graduate Studies in the Department of Electrical and Computer Engineering at the University of Illinois at Chicago.
In 2009, Khokhar was named a Fellow of the Institute of Electrical and Electronics Engineers (IEEE). His accolades also include the NSF Career Award and several best paper awards.
Khokhar's research focuses on context-aware wireless networks, computational biology, machine learning in healthcare, content-based multimedia modeling, retrieval, multimedia communication, and high-performance algorithms. He is renowned for his expertise in developing high-performance solutions for data- and communication-intensive multimedia applications. With over 360 papers published in peer-reviewed journals and conferences, his research has garnered support from the National Science Foundation, the National Institutes of Health, the United States Army, the Department of Homeland Security, and the Air Force Office of Scientific Research.
Khokhar holds a bachelor's degree in electrical engineering from the University of Engineering and Technology in Lahore, Pakistan, a master's degree in computer engineering from Syracuse University, and a Ph.D. in computer engineering from the University of Southern California.
\end{IEEEbiography}

\vfill\pagebreak

\end{document}